\documentclass{article}

\usepackage[preprint]{neurips_2025}

\usepackage[utf8]{inputenc} 
\usepackage[T1]{fontenc}    
\usepackage{hyperref}       
\usepackage{url}            
\usepackage{booktabs}       
\usepackage{amsfonts}       
\usepackage{amsmath}
\usepackage{amssymb}

\usepackage{mathtools}
\usepackage{amsthm}
\usepackage{multirow}
\usepackage{makecell}
\usepackage{nicefrac}       
\usepackage{microtype}      
\usepackage{xcolor}         

\usepackage{floatflt}
\usepackage{wrapfig,booktabs}
\usepackage{graphicx}
\usepackage{nicefrac}
\usepackage{algorithm}
\usepackage{algpseudocode}

\theoremstyle{plain}

\theoremstyle{definition}

\theoremstyle{remark}

\newcommand{\ourmethod}{iw-SFT}

\usepackage[textsize=tiny]{todonotes}

\title{Supervised Fine Tuning on Curated Data is Reinforcement Learning (and can be improved)}

\author{%
  Chongli Qin \thanks{Correspondence should be addressed to: contact@independentresearch.ai. Code and further information can be found at \url{https://independentresearch.ai/posts/iwsft}.} \\
  Independent Research\\
  San Francisco, USA \\
  \And
  Jost Tobias Springenberg $^*$ \\
  Independent Research \\
  San Francisco, USA \\
}

\begin{document}
\maketitle

\begin{abstract}
Behavior Cloning (BC) on curated (or filtered) data is the predominant paradigm for supervised fine-tuning (SFT) of large language models; as well as for imitation learning of control policies.
Here, we draw on a connection between this successful strategy and the theory and practice of finding optimal policies via Reinforcement Learning (RL). 
Building on existing literature, we clarify that SFT can be understood as maximizing a lower bound on the RL objective in a sparse reward setting. Giving support to its often observed good performance. From this viewpoint, we realize that a small modification to SFT leads to an importance weighted variant that behaves closer to training with RL as it: i) optimizes a tighter bound to the RL objective and, ii) can improve performance compared to SFT on curated data.
We refer to this variant as importance weighted supervised fine-tuning (\ourmethod). We show that it is easy to implement and can be further generalized to training with quality scored data. The resulting SFT variants are competitive with more advanced RL algorithms for large language models and for training policies in continuous control tasks. For example achieving 66.7\% on the AIME 2024 dataset.
\end{abstract}

\section{Introduction}
Post-training of large language models (LLMs) is an increasingly important component in modern large model training pipelines since it can help bring out latent capabilities dormant in pre-trained LLMs. Whether post-training should be performed via reinforcement learning (RL) or instead use simple supervised finetuning (SFT), and how these two steps can be combined, is a hot topic in the community. Generally, Reinforcement learning (RL) from human~\citep{ouyang2022training, ziegler2019fine} or AI feedback \citep{lee2023rlaif} has become a crucial step in aligning large language models (LLMs) to better follow human intent, such as instructions, or to steer models to generate outputs that are more helpful and factual~\citep{bai2022constitutional}. Recently, it has been established that RL is an essential part for eliciting reasoning capabilities in LLMs as exeplified by the recent OpenAI o1 and DeepSeek-R1~\citep{guo2025deepseek} models. 

Unfortunately, RL is difficult to interpret and tune. As a result there is a large body of research aiming to further stabilize or simplify existing RL algorithms (see e.g.~\citet{shao2024deepseekmath,ahmadian2024back,li2023remax}). Many recent post-training approaches even attempt to side-step RL training entirely -- e.g. by directly training LLMs to maximize preferences such as in DPO~\cite{rafailov2024direct} or other RL-free approaches~\citep{abdolmaleki2024preference}. Intriguingly, some recent work~\citep{muennighoff2025s1,SingleSampleRL} goes as far as showing that one can improve reasoning in LLMs via SFT on a carefully filtered/curated dataset and then scaling inference compute.

In this paper, we draw a on a known connection between RL as inference~\citep{PetersRWR,KoberPower,LevineEM} and (iterative) maximum likelihood supervised learning that we believe is underexplored. As we show, this perspective suggests that when we perform SFT on filtered data we are in fact optimizing for a loose lower bound on the RL objective in a sparse reward setting (induced by the filtering of the data). By taking a fresh look at SFT from this perspective, we show that this bound can be tightened significantly in two ways: by sampling data proportional to quality scores where they are available,  and via importance re-weighting of the maximum likelihood SFT objective. Intuitively, this re-weighting can also be seen as a way of performing `adaptive-filtering' since the reweighted data points should place increasingly heavier weights on 'preferred' data points. The resulting small modifications to SFT -- which we refer to as SFT from quality sampled data (SFT(Q)) and importance weighted SFT (\ourmethod) respectively -- are easy to implement and can match or surpass SFT in many settings. Our experiments show that when optimizing with \ourmethod\ on curated data, reasoning naturally emerges in LLMs with no need to further scale inference compute (e.g. via some form of budget forcing~\citep{muennighoff2025s1}); leading to high performance on benchmarks such as AIME (66.7\%) or GPQA (64.1\%) outperforming standard SFT on the same data. We further show the generality by adapting our method to train continuous control policies in an offline RL setting.
Our contributions are as follows:
\begin{itemize}
    \item We demonstrate that supervised fine tuning is optimizing a lower bound on the RL objective in the sparse reward setting and introduce an improved, weighted, variant of SFT, \ourmethod. 
    \item We perform extensive experiments, including training LLMs for reasoning (evaluating on AIME 2024 and MATH500~\citep{hendrycks2021measuring}), outperforming the state-of-the-art for training open-models on open data~\cite{muennighoff2025s1}. We also show how \ourmethod\ can improve upon BC in offline RL for control on the D4RL benchmark~\citep{fu2020d4rl}. 
\end{itemize}

\section{Related Work}

Our work connects several different fields of study which are all closely related. Several works have previously considered connections between supervised fine-tuning and RL. For example, methods from the literature on reward weighted regression~\citep{DayanEM,PetersRWR} (RWR) for RL have been adopted to improve supervised learning in cases where reward functions are available; e.g. in the literature on iterative maximum likelihood~\citep{NiIWPMRG23,AgarwalSparse,Norouzi}. Most related to our approach, recent work has considered  reward weighted iterative filtering (or rejection sampling) for supervised fine-tuning~\citep{yuan2024scaling,durwsft,gulcehrerest}, showing success in fine-tuning LLMs. Among them the ReST family of methods~\citep{gulcehrerest,singh2024beyond} can be understood as iteratively optimizing the SFT bound we discuss below (though without importance weighting). 
In a related direction, recently, DeepSeekMath~\citep{shao2024deepseekmath} also drew on the connection between expectation maximization, RWR and supervised learning, treating methods such SFT, RFT, DPO, PPO, GRPO as maximizing the obective:
$
\mathbb{E}_{(q, o)\sim\mathcal{D}}\left( \frac{1}{|o|}\sum^{|o|}G(q, o, r) \nabla_{\theta}\log \pi_\theta(o_t|q, o_{<t})\right),
$ where $G$ denotes generalized returns.

In this work, we build on a similar connection between SFT and RL. In particular, we realize that by following prior work on RL as inference (see e.g. \citet{DayanEM,PetersRWR,PetersREPS,LevineEM}) we can interpret SFT as optimizing a lower bound to the RL objective. 
In particular we build on a bound that is a special case of a general form developed in~\citet{RouxPower}. In this work, the authors show that the well known PoWER \citep{KoberPower} and reward weighted regression (RWR) \citep{DayanEM,PetersRWR} algorithms can be derived by finding a lower bound to the RL objective, and can be generalized to an iterative scheme. Follow-up work \citep{roux2016tighter} also considered the applicability of this loss to supervised classification, albeit in a different way than what we consider here. The authors consider introducing an iterated supervised learning scheme which leads to optimizing a tighter bound to the classification accuracy than the commonly used negative log likelihood loss. In contrast, we here are concerned with fine-tuning models on curated (or filtered) datasets that were originally drawn from some reference distribution. 

In the literature one can also find general discussions on training generative models on self-generated data after human curation e.g. in~\citet{palms,selfconsuming}. Most relevant to the ideas presented here is recent work on interpreting generative models that consume their own, human curated, data as optimizing human preferences via RLHF \citep{selfconsuming}. This work again invokes the well known connections between SFT and EM based policy search.

Our connection between RL and SFT is focused on the sparse reward regime. As a result, it is closely related to other works that aim to improve training on preference data. This includes different RLHF methods such as \citet{ouyang2022training, ziegler2019fine}. As well as methods which bypass explicit RL such as DPO~\citep{rafailov2024direct} or expectation maximization~\citep{abdolmaleki2024preference}.

In the literature on RL for control, many recent works have explored the connection between expectation maximization and KL regularized policy optimization~\citep{abdolmaleki2018maximum,MPOoffline,AWRPeng19,NairAWAC,Schulman17}. Similar to these, our methodology can be used both in the off-policy or online setting. In this work, we mostly focus on the offline setting. For our experimentation on MuJoCo Locomotion tasks, we compare with highly performant offline RL algorithms such as AWAC~\citep{NairAWAC} and IQL~\citep{kostrikov2021offline} as well as top-K percentile based behavior cloning (BC)~\citep{Brandfonbrenner,ChenDT}.

\section{SFT on Curated Data as Reinforcement Learning}
Reinforcement Learning (RL) is the predominant approach for improving alignment and reasoning in LLMs; as well as for training agents in (continuous) control settings. RL, however, is often compute and data intensive and hard to tune. In contrast, SFT on curated (or filtered) datasets is stable, requires no change in the loss-function between pre-training and fine-tuning and can be surprisingly effective.

Here, we make a connection between these two algorithms: SFT and RL. Specifically, we show that SFT on filtered data optimizes a lower bound on the RL objective for a sparse reward task induced by the curation/filtering process. From this perspective we can observe that this lower bound becomes looser as the model moves further away from the expert that generated the data (the reference policy). Interestingly, following existing literature on bounding the RL objective~\citep{RouxPower} we can show that one can introduce an adaptive (importance) reweighting scheme (\ourmethod) that asymptotically maximizes the RL objective as the bound tightens over the course of training. As we will see this simple change from SFT to \ourmethod{} not only theoretically narrows the gap between SFT and RL but also practically can achieve better results in the same settings that SFT excels in.

\subsection{Notation and Background}
To connect RL and SFT we formalize the general RL problem and then show how SFT can be seen as maximizing a lower bound on this objective; specializing a well known bound from the RL literature \citep{PetersRWR, PetersREPS, roux2016tighter} to the sparse / binary reward setting. 
\paragraph{Notation}In the following, $s \in \mathcal{S}$ and $a \in \mathcal{A}$ denote states and actions in a Markov Decision Process (MDP) respectively. We define a trajectory as a sequence of states and actions: $\tau = (s_0, a_0, \cdots, s_T) \in \mathcal{S} \times \mathcal{A} \cdots \mathcal{S} = \mathcal{B}$. The actions are drawn from a parametric policy $\pi(a_t | s_t, \theta)$ (a probability distribution over actions) and the dynamics of the trajectory through time depend on the transition probabilities $p(s_{t+1} | s_t, a_t)$. The reward function is dependent on the state and action $r(s_t, a_t)$, we simplify this to $r(t)$; where $r(T)$ then corresponds to the terminal reward. We aim to maximize the discounted cumulative reward or the return, $R(\tau) = \sum_{t=0}^T \gamma(t) r(t)$, with discount factor $\gamma$. Note that in a sparse reward setting where we only observe a binary terminal reward $r(T)$ and assume  $\forall t \neq T: r(t) = 0$  and use $\gamma = 1$ the return simplifies to $\mathbb{I}(R(\tau)) = \sum_{t=0}^T r(t) = r(T)$ (where $\mathbb{I}$ denotes an indicator function). Often such a sparse reward setting is induced by checking if a continuous function exceeds a threshold at the last timestep. 

\paragraph{Reinforcement learning}
Using the above, we can define the probability of a given trajectory as:
\begin{align}
p_{\pi}(\tau; \theta) &= p(s_0)\prod_{t=0}^{T
- 1}p(s_{t + 1} | s_t, a_t) \pi(a_t | s_t; \theta).
\end{align}
We further note that, in the language modeling setting the states $s$ and actions $a$ are both given by the tokens emitted by the model -- i.e. $\mathcal{S} = \mathcal{A}$ -- and the transition dynamics are deterministic. Thus, the policy directly determines the next state and we can equivalently write $p(\tau; \theta) = p(s_0)\prod_{t=0}^{T - 1} \pi(s_{t+1} | s_0, \cdots, s_t; \theta)$, which is the autoregressive model probability of the text tokens. Note that in this case the transitions are not Markovian and we thus condition on all previous states but all derivations below apply without loss of generality in both cases.

The objective of training a policy $\pi$ via reinforcement learning (RL) is defined as maximizing the expected cumulative return, where the expectation is over observed trajectories:
\begin{align}
\mathrm{max}_{\theta}~J(\theta) = \int_{\mathcal{B}} p(\tau; \theta) R(\tau) \mathrm{d}\tau \label{eq:rl_objective} = \mathbb{E}_{p(\tau; \theta)}\big[R(\tau)\big].
\end{align}

To see the connection between RL and SFT let us first assume that we only have access to trajectories $\tau \in \pi_{\mathrm{ref}}$ generated by some reference policy. This could be a human demonstrating desired trajectories/outputs/behaviour or simply a pre-trained model that we query for outputs.
To use these trajectories for optimizing the RL objective we have to re-express it in terms of these samples. Here we need to make the assumption that the support of $p_{\pi}$ is a subset of $\pi_{\mathrm{ref}}$. Using importance sampling (see ~\citep{Kahn1953MethodsOR,Rubinstein,ripley2009stochastic} for a review), we can then express our RL objective from Eq.~\ref{eq:rl_objective} as:
\begin{align}
    J(\theta) &= \int_{\mathcal{B}} \pi_{\mathrm{ref}}(\tau) \frac{p_{\pi}(\tau; \theta)}{\pi_{\mathrm{ref}}(\tau)} R(\tau) \mathrm{d}\tau = \mathbb{E}_{\pi_{\mathrm{ref}}(\tau)}\Big[ \frac{p_{\pi}(\tau; \theta)}{\pi_{\mathrm{ref}}(\tau)} R(\tau)\Big],
\end{align}
we overload the notation of $\pi_{\text{ref}}$ for simplicity, where the expectation is over the reference $\pi_{\mathrm{ref}}(\tau) = p(s_0)\prod_{t=0}^{T
- 1}p(s_{t + 1} | s_t, a_t) \pi_\mathrm{ref}(a_t | s_t)$ and weight samples via the importance weight $\nicefrac{p_{\pi}(\tau; \theta)}{\pi_{\mathrm{ref}}(\tau)}$.

\subsection{SFT lower bounds RL } 
We can derive a lower bound on this RL objective via a well known inequality \citep{KoberPower,RouxPower}. Assuming strictly positive rewards, we can utilize the inequality $x \geq 1 + \log(x)$ applied to the importance weights, to show that (for more details see Appendix~\ref{Appendix:Derivation}):
\begin{align}
J(\theta) \geq  \int_{\mathcal{B}}\pi_{\mathrm{ref}}(\tau) \log  p(\tau; \theta)R(\tau) \mathrm{d}\tau + \mathrm{cst}\label{eq:loose_bound} = \mathbb{E}_{\pi_{\mathrm{ref}}(\tau)}\Big[R(\tau) \log p(\tau; \theta)\Big] + \mathrm{cst},
\end{align}
where $\mathrm{cst}$ denotes constant terms independent of $\theta$. This is a well known objective in the RL literature \citep{DayanEM,PetersRWR,KoberPower,RouxPower} often referred to as reward weighted regression. We can relate this objective to supervised learning in two cases (i) we assume access to a dataset of filtered ``good data'' (e.g. a natural setting in cases where we have success indicators assigned by humans); (ii) we additionally assume access to quality scores for all ``good data'' (i.e. we assume each $\tau$ can be given a quality score on an ordinal scale).
For case (i) we can find a relation by assuming binary, sparse, rewards in the RL objective. Then the return simplifies to an indicator function, $R(\tau) = \mathbb{I}(S(\tau) > 0)$ (where $S$ is some score function indicating success), collapsing Equation~\ref{eq:loose_bound} to the maximum likelihood supervised fine tuning objective:
\begin{align}
   J(\theta) \geq \mathbb{E}_{\pi_{\mathrm{ref}}(\tau)}\Big[\mathbb{I}(S(\tau) > 0) \log p(\tau; \theta)\Big] = c_{\mathrm{ref}}\underbrace{\mathbb{E}_{\tau \in \mathcal{D}^+}\Big[\log p(\tau; \theta)\Big]}_{\mathcal{J}_\text{SFT}(\theta)}, 
    \label{eq:sft}
\end{align}
where $\mathcal{D}^+ = \lbrace \tau_i | \tau_i \sim \pi_{\mathrm{ref}}(\tau), \mathbb{I}(S(\tau_i) > 0) \rbrace_{i=1}^N$ is a dataset of trajectories sampled from $\pi_\textrm{ref}$ and filtered by the binary reward indicator and $c_{\mathrm{ref}} = \mathbb{E}_{\pi_{\mathrm{ref}}(\tau)}[\mathbb{I}(S(\tau) > 0)]$ is constant independent of $\theta$. And we start optimization of $J(\theta)$ by initializing to the reference distribution, setting $\pi_{\theta_0} = \pi_{\mathrm{ref}}(\tau)$. If $\pi_{\mathrm{ref}}$ is not available in closed form we approximate it by first learning it via standard behavior cloning on the full dataset $\mathcal{D}$. 
Thus \emph{we can interpret SFT on filtered data as maximizing a lower bound on the RL objective}.

Analogously, for case (ii) we can assume rewards $R(\tau) = \sum_{i=0}^C \mathbb{I}(S(\tau) > c_i)$, with $C$ (ordinal) quality classes. Then RL simplifies to maximum likelihood of trajectories proportional to quality:
\begin{align}
    J(\theta) \geq \mathbb{E}_{\pi_{\mathrm{ref}}(\tau)}\Big[\sum_{i=0}^C\mathbb{I}(S(\tau) > c_i) \log p(\tau; \theta)\Big] =c_{\mathrm{ref}}^Q \underbrace{\mathbb{E}_{\tau \in \mathcal{D}_Q^+}\Big[\log p(\tau; \theta)\Big]}_{\mathcal{J}_\text{SFT(Q)}(\theta)}, 
    \label{eq:sftq}
\end{align}
where $\mathcal{D}_Q^+ = \mathcal{D}_{c_1}^+ \cup \dots \cup \mathcal{D}_{c_C}^+$ with $\mathcal{D}_c^+ = \lbrace \tau_i | \tau_i \sim \pi_{\mathrm{ref}}(\tau), \mathbb{I}(S(\tau_i) > c_i) \rbrace_{i=1}^N$. Note $c_{\mathrm{ref}}^Q = \mathbb{E}_{\pi_{\mathrm{ref}}(\tau)}[\sum_{i=0}^C\mathbb{I}(S(\tau) > c_i) ]$. And we have the relation $\forall i: c_i > c_{i-1}$. Hence, we can interpret SFT over quality curated data as maximizing a lower bound on the RL objective. 

It is important to realize, however, that both lower bounds, become looser throughout training, as the distance between $p(\tau; \theta)$ and $\pi_{\mathrm{ref}}(\tau)$ increases. 
This connection between RL via reward weighted regression and maximum likelihood is not new \citep{RouxPower,roux2016tighter,selfconsuming} but we believe it is underexplored in the context of SFT and RL for large models. 

\subsection{Improving SFT from the perspective of RL}
The bound SFT optimizes becomes looser as we train, thus weakening the connection to RL. We show that by re-framing the RL problem we can introduce a tighter bound that only requires a minimal change to the maximum likelihood objective used in SFT and can achieve better results.

To derive this variant we first introduce an auxiliary distribution $q(\tau)$ that we are free to choose. And we will choose it later to trade off the tension between staying close to the reference policy for stable optimization and tightening the bound. Using this auxiliary we can re-write the RL objective as:
\begin{align}
    J(\theta) = \int_{\mathcal{B}} \pi_{\mathrm{ref}}(\tau)\frac{q(\tau)}{\pi_{\mathrm{ref}}(\tau)}  \frac{p(\tau; \theta)}{q(\tau)} R(\tau) \mathrm{d}\tau.
\end{align}
By applying the inequality $x \geq 1 + \log(x)$ to $x = \nicefrac{p(\tau; \theta)}{q(\tau)}$, as previously considered in \citep{roux2016tighter,KoberPower} we can obtain an analogous bound to \eqref{eq:loose_bound}: 
\begin{align*}
    J(\theta) & \geq \int_{\mathcal{B}} \pi_{\mathrm{ref}}(\tau) \frac{q(\tau)}{\pi_{\mathrm{ref}}(\tau)} \log p(\tau; \theta) R(\tau) \mathrm{d}\tau + \mathrm{cst}  = \mathbb{E}_{\pi_{\mathrm{ref}}(\tau)}\Big[\frac{q(\tau)}{\pi_{\mathrm{ref}}(\tau)} R(\tau) \log p(\tau; \theta)\Big] + \mathrm{cst}.
\end{align*}
For simplicity we drop the constant (wlog). If we assume sparse rewards as above we can further turn this into a training objective over filtered data that results in an importance weighted version of SFT:
\begin{align}
    J(\theta) & \geq \mathbb{E}_{\pi_{\mathrm{ref}}(\tau)}\Big[\frac{q(\tau)}{\pi_{\mathrm{ref}}(\tau)} \mathbb{I}(S(\tau)) \log p(\tau; \theta)\Big] = c_{\mathrm{ref}}\underbrace{\mathbb{E}_{\tau \in \mathcal{D}^+}\Big[\frac{q(\tau)}{\pi_{\mathrm{ref}}(\tau)} \log p(\tau; \theta)\Big]}_{\mathcal{J}_{\text{\ourmethod}}},
    \label{eq:iwsft}
\end{align}
where $\mathcal{D}^+ = \lbrace \tau_i | \tau_i \sim \pi_{\mathrm{ref}}(\tau), \mathbb{I}(S(\tau_i)) > 0 \rbrace_{i=1}^N$ and $c_{\mathrm{ref}} = \mathbb{E}_{\pi_{\mathrm{ref}}(\tau)}[\mathbb{I}(S(\tau) > 0)]$. Here, the auxiliary distribution $q(\tau)$ is now used to importance weight the examples in the dataset. And an analogous bound for quality scored data $\mathcal{D}_Q^+ = \mathcal{D}_1^+ \cup \dots \cup \mathcal{D}_C^+$ as in Eq. \eqref{eq:sftq} can be found: 
\begin{align}
J(\theta) & \geq c_{\mathrm{ref}}^Q\mathcal{J}_{\text{\ourmethod(Q)}} = \mathbb{E}_{\tau \in \mathcal{D}_Q^+}\Big[\frac{q(\tau)}{\pi_{\mathrm{ref}}(\tau)} \log p(\tau; \theta)\Big].
\end{align}
Critically, both objectives have the property that as $q(\tau) \rightarrow p(\tau; \theta)$ we have $\mathcal{J}_\text{\ourmethod}(\theta) \rightarrow J(\theta)$. Namely, as $q$ gets closer to our policy distribution, the bound tightens. However, at the same time, tightening the bound will increase variance in the importance weights and it is thus not generally safe to just set $q(\tau) = p(\tau, \theta)$. That being said, and in contrast to standard importance weighting for policy gradients \citep{jiang16,Metelli}, we are free to choose $q$ to smooth out the bound. We can thus side-step common problems with large variance in the importance weights/blow-up by enforcing importance weights to not be too extreme. 
Optimizing $\mathcal{J}_{\text{iwSFT}}$ or $\mathcal{J}_{\text{iwSFT(Q)}}$ then can be achieved by sampling from the filtered data distribution and computing importance weights on the fly (requiring one extra reference model to be kept in memory) as shown in Algorithm \ref{alg:cap} in the appendix.

\subsection{The importance of choosing $q(\tau)$} 
As mentioned, we are free to choose $q(\tau)$ in our method. We would generally like to achieve $q(\tau) \rightarrow p(\tau; \theta)$ over the course of training so that the bound becomes tight. However, as $q(\tau)$ moves away from $\pi_\textrm{ref}$ the variance of the bound (and thus the variance of its gradient) increases. As a result we want to bound the KL divergence $KL(q(\tau) \| \pi_\textrm{ref}(\tau)) < \beta$ but need to trade this off with tightening the bound; a well known problem when dealing with importance sampling based objectives.

In contrast to standard off-policy policy gradient algorithms, choosing $q(\tau)$ such that it strikes this trade-off (e.g. via clipping or other means as discussed below) will not bias our algorithm--but may nonetheless lead to different optima and thus better (or worse) policies.
In practice, we choose
\begin{align}
q(\tau) = p(s_0)\prod_{t=0}^{T
- 1}p(s_{t + 1} | s_t, a_t) \pi(a_t | s_t; \theta_q)
\end{align}
where $\theta_q \rightarrow \theta$ towards the end of training such that $q(\tau) \rightarrow p_{\pi}(\tau; \theta)$.

For controlling the variance, we consider two different variants of bounding the difference between $q(\tau)$ and $\pi_\textrm{ref}$ by controlling the importance weights. In either variant we set $\theta_q$ to be a time-lagged version of $\theta$, the parameters we are optimizing (see experiments section for details) and then constrain the importance weights to control the evolution of $p(\tau, \theta)$ during training (and hence indirectly $q(\theta)$).  We note that for each variant we always consider \textit{importance weights over full trajectories}.

\paragraph{Clipping per step importance weights} A simple way to control the importance weights is to simply bound the per step importance weights to be within a certain range. That is we choose: $q(\tau) = p(s_0) \prod_{t=0}^{T-1}(p(s_{t+1} | s_t, a_t) \text{clip}(\pi(a_t | s_t; \theta), \alpha_\text{min}, \alpha_\text{max})$, where we typically set $\alpha_\text{min}$ and $\alpha_\text{max}$ fixed, i.e. in the language model experiments we will choose $(\alpha_\text{min}, \alpha_{\max}) = (1 - 0.8, 1 + 0.8)$. We will then do an overall clipping of the importance weight: $\text{clip}(q(\tau) / \pi_{\text{ref}}(\tau), \beta_{\min}, \beta_{\max})$.

\paragraph{Smoothing importance weights} As a second, more general, option we consider ensuring low variance by smoothing the importance weights on the trajectory level. In particular we consider transforming the weights as:
\begin{align}
\frac{q(\tau)}{\pi_\mathrm{ref}(\tau)}  = \exp \left(\sum_i g(\rho_i) \right),\quad \rho_i = \log(\pi_q(a_i | s_i)) - \log(\pi_{\mathrm{ref}}(a_i | s_i))
\end{align}
Here, $i$ denotes the $i$-th token in the sequence $\tau$, and $g(x)$ is a monotonically increasing function with respect to $x$. We use a simple temperature function for continuous control experiments\footnote{But the above clipping can also be expressed using this formulation as $g(x) = \mathrm{clip}(x, \min, \max)$.}: $g(x) = kx$, where $k$ is a hyperparameter, for all our experiments in continuous control domains. By setting $k = \nicefrac{1} / {\sum_i M_i}$ (where $M$ is the token mask for a transformer LLM) we can implement using the averaged log importance weight. We instead use $k = \alpha \sum_i M_i$ which gives us an extra degree of freedom to smooth out the importance weights. This way we can smoothly move from no importance weighting ($\alpha \rightarrow 0$) to more and more ``peaked'' weights. We ablate the choice of $\alpha$ in the supplementary material.

\subsection{A Toy Example to build Intuition for SFT on curated/filtered data}

We consider a toy example to get an intuition for when the bound optimized via SFT is problematic--and thus when the importance weighted version can be beneficial. We construct this toy example to answer the following questions: \textbf{Q1:} Does the loose lower bound lead to suboptimal policies? \textbf{Q2:} What is the effect of the importance weighting? Can it help learn better policies? \textbf{Q3:}  What are the failure modes of the importance weighted variant, and how can they be avoided?

Consider a simple bandit optimization problem in which an agent has a choice of two actions $\mathcal{A} = \lbrace \text{pull-left}, \text{pull-right} \rbrace$. The episode terminates after picking a single action. $R(\tau)$ is either 0 or 1; the probability distribution of $R(\tau)$ is defined by:
\begin{align}
    \mathbb{P}(R(\tau)=1) = \left\lbrace \begin{array}{l l}
    0.5 & \text{if } \tau = \text{pull-left} \\ 1 & \text{if } \tau = \text{pull-right}
    \end{array} \right.
\end{align}
\begin{wrapfigure}{r}{0.35\textwidth}
\vspace{-0.45cm}
    \centering
    \includegraphics[width=0.35\textwidth]{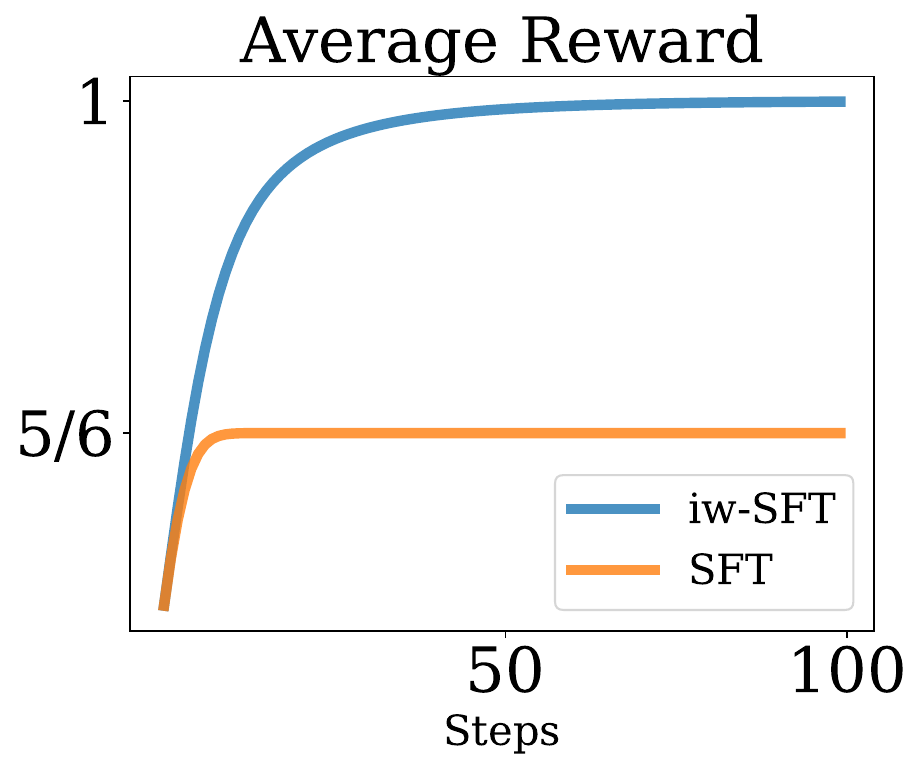}\caption{\small{Toy example showing a failure mode of SFT on curated data.}}\label{fig:failuremode}\vspace{-0.45cm}
\end{wrapfigure}
Thus the optimal strategy is to always pull the right arm. We assume that we can observe the behavior of the system only by watching (collecting data from) a reference policy $\pi_{\mathrm{ref}} (a) = \mathcal{U}\lbrace \text{pull-left}, \text{pull-right}  \rbrace$ that picks between the two arms uniformly at random. 

We collect a dataset $\mathcal{D}$ using this policy which will contain $50\%$ pull-left and $50\%$ pull-right actions. 
Filtering this dataset by the binary success reward to produce $\mathcal{D}^+ = \lbrace \tau_i | \tau_i \sim \pi_{\mathrm{ref}}(\tau), R(\tau_i) = 1 \rbrace_{i=1}^N$ then results in a dataset with 2 times more pull-right than pull-left actions. 
Subsequently training via SFT on this data results in a policy $\pi_\text{SFT}$ that chooses pull-left one third of the time and pull-right two thirds of the time. The resulting average reward is $5/6$. 
See Figure~\ref{fig:failuremode} for a numerical experiment. This policy improves on the reference policy, but is also not optimal. The key problem here is that the SFT procedure cannot incorporate knowledge from failures. For \textbf{Q1}: while SFT on curated data can be seen as optimizing a lower bound on the RL objective and will improve on the reference policy it can result in sub-optimal behavior.

Suppose we introduce importance weights (according to Equation~\eqref{eq:iwsft}) where $q(\tau) = \pi(\tau; \theta)$.
Now over the course of training, we will adaptively put more weight on pull-right compared to pull-left actions until we recover a policy that will pull-right every time. 
This can also be seen in Figure~\ref{fig:failuremode}. \emph{Thus importance weighting can recover information from failures}. Indeed, if we train with \ourmethod~on this data, we recover the optimal policy of just choosing pull-right; answering \textbf{Q2}: \ourmethod~can ``recover'' information from discarded samples. 

Finally regarding the failure modes (\textbf{Q3}): while we recover the optimal policy with \ourmethod~in this case, it does lead to collapse of the policy to a single action (which happens to be correct here). In general, the importance weights determine how far away from the reference policy we allow $\pi_\theta$ to deviate. If we place no bound on the importance weights one can easily see that in a high-dimensional case, they might grow out of bounds (the infinite variance problem \citep{ISvar}) or lead to us discarding too much of the filtered data. 
As mentioned above, this can be mitigated by forcing $q(\tau)$ to stay close to $\pi_{\mathrm{ref}}$--e.g. via clipping.

\section{Experiments}
We evaluate our proposed improvements to SFT, and compare to existing RL algorithms in two settings: 1) training LLMs for reasoning, 2) training control policies from offline data. 

\subsection{Training LLMs for reasoning via SFT on curated data}
For our initial set of experiments we consider fine-tuning an open-source LLM (Qwen2.5-32B-Instruct \citep{yang2024qwen2}) to improve its reasoning capabilities for challenging maths problems.
To finetune the model, we use the S1.1K dataset introduced in \citet{muennighoff2025s1}. This is a heavily curated dataset consisting of $1,000$ high quality reasoning traces. These examples were created by first collecting a set of 59K candidate question/answer pairs from existing datasets~\citep{numina_math_datasets,aime,gao2024omnimathuniversalolympiadlevel,zhong2023agievalhumancentricbenchmarkevaluating} and generating reasoning traces and responses for the respective questions using Gemini Flash \citep{geminithinking}. The dataset is then decontaminated against the test datasets and filtered down to 1k examples by quality and diversity as described in \citet{muennighoff2025s1}. This is thus a great example of a high-quality, curated dataset.

\begin{table}[tb]
\caption{Results for AIME2024, MATH500, GPQA Diamond compared to existing baselines. Most baseline results are taken from \citep{muennighoff2025s1}. We include comparisons to: \citep{yang2024qwen2,qwq-32b-preview,o1,guo2025deepseek,bespoke_stratos,sky_t1, geminithinking, gemini2_5}. }\label{tab:perf}
\centering
\begin{tabular}{lrrrr}
\toprule
Model & \makecell{\# ex.} & \makecell{AIME 2024} & \makecell{MATH 500} & \makecell{GPQA Diamond} \\
\midrule
\multicolumn{5}{c}{\textbf{API only}} \\
\midrule
o1-mini & N.A. & 70.0 & 90.0 & 60.0 \\
o1 & N.A. & \textbf{74.4} & \textbf{94.8} & \textbf{77.3} \\
Gemini 2.0 Flash Think.& N.A. & 60.0 & N.A. & N.A. \\
Gemini 2.5 & N.A. & \textbf{92.0} & N.A. & N.A. \\
\midrule
\multicolumn{5}{c}{\textbf{Open Weights}} \\
\midrule
Qwen2.5-32B-Instruct  & N.A. & 26.7 & 84.0 & 49.0\\
QwQ-32B & N.A. & 50.0 & 90.6 & 54.5 \\
r1 & $\gg$800K & \textbf{79.8} & \textbf{97.3} & \textbf{71.5} \\
r1-distill & 800K & 72.6 & 94.3 & 62.1 \\
\midrule
\multicolumn{5}{c}{\textbf{Open Weights and Open Data}} \\
\midrule
Sky-T1 & 17K & 43.3 & 82.4 & 56.8 \\
Bespoke-32B & 17K & \textbf{63.3} & 93.0 & 58.1 \\
\midrule
s1 w/o BF & \textbf{1K} & 50.0 & 92.6 & 56.6 \\
s1 with BF "Wait" 4x & \textbf{1K} & 56.7 & 93.0 & 59.6 \\
s1.1 w/o BF & \textbf{1K} & 56.7 & 94.4 & 60.6 \\
s1.1 with BF "Wait" 2x &\textbf{1K} & 56.7 & \textbf{95.4} & \textbf{63.6} \\
\ourmethod (per-step iw) & \textbf{1K} & \textbf{63.3} & \textbf{95.2} & 60.6 \\
\ourmethod& \textbf{1K} & \textbf{66.7} & 94.8 & \textbf{64.1} \\
\bottomrule
\end{tabular}
\vspace{-1em}
\end{table}

The hyperparameters of our model are set the same as in \citep{muennighoff2025s1} (we use cosine annealing for the learning rate and AdamW~\citep{loshchilov2017fixing}) details are given in the appendix. We also use $g(x) = 0.1 \times \mathrm{clip}(x, 0.2, 1.8)$ to smooth/clip the importance weights. Given that we do not have access to the model weights for Gemini Flash (which was used to generate the reasoning traces), or its original training data, we also have to rely on an approximation for the reference distribution $\pi_\mathrm{ref}$ in the importance weights. Ideally, we would pick a model that is ``close'' to Gemini-Flash as an approximation. We tested multiple models at the 32B parameter level, comparing their summed log-probabilities over examples from the dataset and find that they assign similar likelihood (albeit giving different top-k next token predictions) and settled on simply using the model we start from as the reference model such that $q, \pi_\mathrm{ref}$ and $\pi_\theta$ are all close at the beginning of training.

All results and comparisons to models from the literature are shown in Table~\ref{tab:perf}. In the table we also include an ablation of our method which uses importance weighting but on the token level rather than sequence level; this result is shown as \ourmethod(per-step iw).
As can be seen from the table, our method outperforms the original s1 results (which also performs SFT, starting from the same checkpoint) on AIME 2024 as well as GPQA Diamond while matching it on MATH 500.  
Interestingly, we find that \ourmethod{} does not require budget forcing (a method introduced in \citet{muennighoff2025s1} to force the model to keep thinking by injecting ``Wait'' tokens) to reach its peak performance. This suggests that the importance weighted objective itself can help extract information from the training data, foregoing the need for extra test time optimization. For ablations please see Appendix~\ref{Appendix:budget}.
The model that only uses per-step importance weights also reaches high-performance across benchmarks, albeit losing 2 percent compared to full sequence importance weighting on GPQA Diamond.

Overall, training with \ourmethod{} on the curated dataset S1.1K introduced by \citet{muennighoff2025s1} surpasses training with SFT only (s1 / s1.1 in the table) by up to 7\% depending on the AIME benchmark. This sets the state-of-the art for open models trained on open data; although better closed models exist as shown in the comparison table. 

\subsection{Offline RL for Continuous Control}
To probe for the general applicability of our approach, we turn to learning policies for continuous control using the offline RL benchmark datasets from D4RL \citep{fu2020d4rl}. We consider the standard locomotion tasks from the benchmark (halfcheetah, hopper, walker2d) both in the setting where data comes from expert policies (Expert (v2) below) and in the setting where data was extracted from a standard RL training run (Med. Replay (v2)) and is thus more diverse.

Since all trajectories in the datasets come with reward information (which we can interpret as unambiguous quality scores), we consider the quality sampling setting. We first filter the data to only contain the top $10\%$ of trajectories and then turn the cumulative episode rewards into ordinal quality annotations by binning them into 3 overlapping bins, see Appendix~\ref{Appendix:stratefied_data} for details. Our model is a simple 3 layer MLP policy as listed in Appendix~\ref{Appendix:Hyper} and instead of updating the importance weighting distribution $q$ at discrete time-points we use exponential weight averaging $\theta_q = \alpha \theta_q + (1 - \alpha) \theta$ during each step of the optimization (this is similar to how target networks in continuous control RL algorithms are often updated \citep{lillicrap2015continuous,fujimoto2021minimalist}). We smooth the importance weights with $g(x) = kx$. We obtain a reference distribution $\pi_\text{ref}$ by first performing simple behavior cloning on all data (e.g. optimizing Equation \eqref{eq:sft} on all data instead of filtered data).

\begin{table}[t]
\caption{Offline RL results for control suite environments from the D4RL suite of tasks. We omit standard deviations for space but mark statistically equivalent results in bold (see supplementary).}
\label{tab:finetuning}
\centering
\scalebox{0.9}{
\begin{tabular}{ l|| c  c  c  c  c   c  || c   c  c}
    \centering
            & BC & BC & AWAC & TD3 & CQL & IQL & SFT & SFT(Q) & \ourmethod(Q) \\
    Dataset/ Method &    & (10\%) &  & +BC &     &     &     &        &               \\
    \midrule
    \multicolumn{10}{c}{\textbf{Medium Replay (v2)}} \\ \midrule
    halfcheetah  & 36.6 & 40.6 & 40.5 & \textbf{44.6} &  \textbf{45.5} & \textbf{44.2} & 35.1 & 39.3 & 40.9 \\
    hopper & 18.1 & 75.9 & 37.2 & 60.9 & \textbf{95.0} & \textbf{94.7} & 79.0 & 84.9 & 85.0 \\
    walker2d & 26.0 & 62.5 & 27.0 & \textbf{81.8} & 77.2 & 73.9 & 58.8 & 66.2 & 75.8 \\
    \midrule
    \multicolumn{10}{c}{\textbf{Expert (v2)}} \\ \midrule
    halfcheetah & 55.2 & \textbf{92.9} & 42.8 & \textbf{90.7} & \textbf{91.6} & 86.7 & \textbf{92.3} & \textbf{92.3} & \textbf{92.6}\\
    hopper & 52.5 & \textbf{110.9} & 55.8 & 98.0 & 105.4 & 91.5 & \textbf{111.2} & \textbf{109.5} & \textbf{111.1} \\
    walker2d & \textbf{107.5} & \textbf{109.0} & 74.5 & \textbf{110.1} & \textbf{108.8} & \textbf{109.6} & \textbf{109.1} & \textbf{109.0} & \textbf{109.0}\\
    \midrule
    \multicolumn{10}{c}{\textbf{Ant-maze (v0)}} \\ \midrule
    umaze & 54.6 & 62.8 & 56.7 & \textbf{87.6} & 70.1 & \textbf{86.7} & 59.7 & 60.7 & 63.1 \\
\end{tabular}}
\end{table}

Results are given in Table \ref{tab:finetuning}. We evaluated SFT, SFT(Q) and \ourmethod(Q) and list results alongside baselines from \citet{kostrikov2021offline} consisting of behavior cloning (BC), which is equivalent to SFT but with parameters initialized randomly. BC on the top-10\% of data (BC 10\%) as well as four state-of-the-art offline RL algorithms (AWAC~\citep{nair2020awac}, TD3+BC~\citep{fujimoto2021minimalist}, CQL~\citep{kumar2020conservative}, IQL~\citep{kostrikov2021offline}). First, we can observe that SFT performs similar to BC (10\%), suggesting that initializing to a policy trained on all data does not lead to meaningful generalization benefits in these tasks. Second, SFT(Q) clearly outperforms BC, BC(10\%) and SFT in all tasks. This shows that training with quality scores, where available, is indeed helpful. Furthermore, while there are algorithms achieving better results in some domains, SFT(Q) is competitive with strong RL baselines in all settings; confirming that the bound optimized by SFT(Q) can meaningfully optimize for the RL objective. Finally iw-SFT(Q) further leads to a slight improvement over SFT(Q), albeit not as large of an improvement as in the LLM case; suggesting that perhaps this data (which stems from trained policies) is too prototypical, making it hard to outperform quality filtering. Overall, we show that our improved SFT variants can be competitive to offline RL for continuous control.

\subsection{Fine-tuning control policies with little data and noisy quality scores}
In a final experiment we test whether fine-tuning can be helpful for control tasks if less prototypical data is available. For this purpose we consider the the Franka kitchen task~\citep{fu2020d4rl}, in which a robot has to interact with a number of items in a simulated kitchen in a specific order. We first pre-train a BC policy on the ``partial'' part of the dataset (containing only some successful trajectories) and then fine-tune on the top $5 \%$ trajectories from the ``complete'' dataset of trajectories (which are all successful but are not enough trajectories to learn the task from) and top $5 \%$ of data from the ``mixed'' part of the dataset (which solve parts of the task but are not fully successful). Results are depicted in Table \ref{tab:kitchen}. As can be seen, BC on the complete data is reasonably successful while training only on the $5 \%$ filtered data fails. Fine-tuning on this data maintains performance, and closes the gap to BC on complete data significantly. Weighting by quality (SFT(Q)) closes the gap further and \ourmethod{}(Q) finally is able to recover performance.

\begin{table}[t]
\caption{Results for Franka Kitchen domain~\citep{fu2020d4rl}.Results are from 100 evaluation episodes after training; averaged over 3 training runs (with standard deviations showing variability over runs). partial $\rightarrow$ complete denotes pre-training on kitchen-partial and fine-tuning on top 5\% from complete.}
\label{tab:kitchen}
\centering
\scalebox{1.}{
\begin{tabular}{ l|| c || c | c  c  c}
    \centering
    Dataset/ Method                & BC (all data)     & BC (5\%)         &  SFT   & SFT(Q) & \ourmethod(Q) \\
    \midrule
    kitchen-partial                & 37.0 $\pm$  4.2   &        -          &   -     &   -     &  -  \\
    partial $\rightarrow$ complete         & \textbf{62.5} $\pm$ 2.0      &  28.9 $\pm$ 8.4   & 45.6 $\pm$ 8.5 & 58.2 $\pm$ 6.2 & \textbf{61.8 $\pm$ 4.1} \\
\end{tabular}}
\vspace{-1em}
\end{table}

\section{Conclusion and Discussion}
In this work, we draw on a connection between reinforcement learning and supervised fine tuning on filtered datasets. From this perspective, we show that SFT is optimizing a lower bound of the RL objective and propose simple modifications that lead to improved variants. 
First, we confirm that SFT with data sampling proportional to quality scores can form a tighter bound on the RL objective for optimizing human preferences. Second, using importance weighting, \ourmethod, can further tighten the bound with respect to the RL objective leading to strong results for training LLMS to reason. We also demonstrate the versatility of the approach by applying our method for control tasks.

There are several limitations of our wok: while we demonstrate improvements over SFT we do so only for fine-tuning in narrow domains, partly restricted by prohibitively large compute costs for large LLM experiments. More extensive evaluations considering a variety of domains for which reasoning is known to improve LLMs could be considered in the future. 
Furthermore both for the language modeling and control domains our results do not bridge the gap to the best existing models in all cases. It remains to be seen whether the approach presented here could reach the same performance given the 'right' data.
Finally, we want to highlight that, as with any method that is used to align ML models to human preferences, we hope our work will help produce more human aligned models, but acknowledge that it could also be used to produce more harmful models (by combining it with adversarial data curation).

\newpage
\bibliography{bibliography}

\newpage
\appendix
\onecolumn


\section{Derivation for the iw-SFT objective}\label{Appendix:Derivation}
For trajectories $\tau = (s_0, a_0, \cdots, s_T) \in \mathcal{B}$ we aim to maximize the expected cumulative reward or the return, $R(\tau) = \sum_{t=0}^T r(t)$, of a given trajectory by the following
\begin{align}
J(\theta) &= \int_{\mathcal{B}} p(\tau; \theta) R(\tau) \mathrm{d}\tau = \mathbb{E}_{p(\tau; \theta)}[R(\tau)]\\
p(\tau; \theta)  &= p(s_0)\prod_{t=0}^{T-1}p(s_{t + 1} | s_t, a_t) \pi(a_t | s_t; \theta).
\end{align}

Let's assume that we have a fixed dataset, $\lbrace\tau_1, \cdots \tau_n\rbrace = \mathcal{D}$ where $\tau \in \pi_{\mathrm{ref}}(\tau)$ where $\pi_{\mathrm{ref}}(\tau) > 0 \quad \forall \quad p(\tau) > 0$ (for example we can assume $\pi_{\mathrm{ref}}$ as Gaussian), thus we can rewrite $J(\theta)$ as
\begin{align}
    J(\theta) = \int_{\mathcal{B}} \pi_{\mathrm{ref}}(\tau) \frac{p(\tau; \theta)}{\pi_{\mathrm{ref}}(\tau)} R(\tau) \mathrm{d}\tau = \mathbb{E}_{\pi_{\mathrm{ref}}(\tau)}\left[\frac{p(\tau; \theta)}{\pi_{\mathrm{ref}}(\tau)} R(\tau) \right]
\end{align}

Using the following inequality: 
\[
x \geq 1 + \log(x),
\]
we can make the observation that:
\begin{align}
    J(\theta) \geq \mathbb{E}_{\pi_{\mathrm{ref}}(\tau)} \left[ \left(1 + \log \frac{p(\tau; \theta)}{\pi_{\mathrm{ref}}(\tau)}\right) R(\tau) \right].
\end{align}

Note that as $\pi_{\mathrm{ref}}$ is fixed; this bound cannot be tightened. Here we can introduce a distribution $q(\tau)$:
\begin{align}
    J(\theta) = \mathbb{E}_{\pi_{\mathrm{ref}}(\tau)}\left[\frac{q(\tau)}{\pi_{\mathrm{ref}}(\tau)}  \frac{p(\tau; \theta)}{q(\tau)} R(\tau) \right].
\end{align}
Which leads to the relation:
\begin{align}
    J(\theta) \geq \mathbb{E}_{\pi_{\mathrm{ref}}(\tau)}\left[ \frac{q(\tau)}{\pi_{\mathrm{ref}}(\tau)} \left(1 + \log \frac{p(\tau; \theta)}{q(\tau)}\right) R(\tau) \right].
\end{align}
Note that in this case if $q(\tau) \rightarrow p(\tau; \theta)$, the bound becomes equality. We note that we can thus optimize the following surrogate objective:
\begin{align}
    \tilde{J}(\theta)  &  = \mathbb{E}_{\pi_{\mathrm{ref}}(\tau)}\left[ \frac{q(\tau)}{\pi_{\mathrm{ref}}(\tau)} \left(1 + \log \frac{p(\tau; \theta)}{q(\tau)}\right) R(\tau) \right] \\
     & =\mathbb{E}_{\pi_{\mathrm{ref}}(\tau)}\left[ \frac{q(\tau)}{\pi_{\mathrm{ref}}(\tau)} \log p(\tau; \theta) R(\tau) \right]+ \mathrm{cst}.
\end{align}
We can drop the constant terms from this objective without loss of generality. From this we can make the following observation where we change our cumulative reward into a filter function; the objective becomes importance weighted maximum likelihood optimization 
\begin{align}
    \tilde{J}(\theta) & = \mathbb{E}_{\pi_{\mathrm{ref}}(\tau)}\left[ \frac{q(\tau)}{\pi_{\mathrm{ref}}(\tau)} \log p(\tau; \theta) \mathbb{I}(R(\tau)) \right] \\
    & \approx \sum_{\tau \in \mathcal{D}^+} \frac{q(\tau)}{\pi_{\mathrm{ref}}(\tau)}  \log p(\tau; \theta)\\ 
    & \approx \sum_{\tau \in \mathcal{D}^+} \frac{1}{n}\sum_{i}^n \frac{q_i(\tau)}{\pi_{\mathrm{ref}}(\tau)}  \log p(\tau; \theta),
\end{align}
note that, here $\mathcal{D}^+$ is the set of data points with positive advantage in our fixed dataset $\mathcal{D}$. When $q = \pi_\mathrm{ref}$, this collapses to maximum likelihood over the positive examples.

\section{Detailed algorithm}
A detailed algorithm listing of our method is given in Algorithm \ref{alg:cap}.
\begin{algorithm}
\caption{\ourmethod / \ourmethod(Q)}\label{alg:cap}
\begin{algorithmic}
\Require $\mathcal{\mathcal{D}}^+$ or $\mathcal{D}_Q^+$ for \ourmethod(Q), $\pi_{\theta_\textrm{ref}}$, $p(\tau; \theta)$, batch size $b$ 
\State $\theta_q \gets \theta_{\mathrm{ref}}$
\State $\theta \gets \theta_{\mathrm{ref}}$
\For{$i\leq N$}
\State Draw $b$ random trajectories $\tau_1, \cdots, \tau_b$ from $\mathcal{\mathcal{D}}^+$ or $\mathcal{D}_Q^+$ respectively
\State $\rho^j_i \gets \log q(\tau_i^j | \tau_{i:i-1}^{j}; \theta_q) -\log \pi(\tau_i^j | \tau_{i:i-1}^{j}; \theta_{\mathrm{ref}})$\Comment{$i$ denotes the $i$th token in $\tau^j$.}
\State $w_j \gets \exp \left(\sum_i g(\rho^j_i)\right)$\Comment{Here $g$ can be temperature function or clipping.}
\State $\Delta \theta \gets \frac{\partial}{\partial \theta} \sum_{j = i_1}^{i_b} w_{j} \log f(\tau^j;\theta)$
\State $\theta \gets \theta - \lambda \mathrm{Optimizer}(\Delta \theta)$
\If{$i \mod Q == 0$}
    \State $\theta_q \gets \theta$\Comment{Updating $q(\theta)$, this update can also be exponential averaging.}
\EndIf
\EndFor
\end{algorithmic}
\vspace{-0.1em}
\end{algorithm}

\section{Hyperparameters for LLM experiments}
The hyperparameters of our model are set the same as in \citep{muennighoff2025s1}: we use cosine annealing for the learning rate, decaying to zero over a maximum of 5 epochs and using a warm-up ratio of 0.05. The target learning rate for AdamW~\citep{loshchilov2017fixing} is $1e-5$,$\beta_1=0.9, \beta_2=0.95$ and the batch size is $8$.  Weight decay is set to $1e-4$. We also use $g(x) = 0.1 \times \mathrm{clip}(x, 0.2, 1.8)$ to smooth/clip the importance weights. Given that we do not have access to the model weights for Gemini Flash (which was used to generate the reasoning traces), or its original training data, we also have to rely on an approximation for the reference distribution $\pi_\mathrm{ref}$ in the importance weights. Ideally, we would pick a model that is ``close'' to Gemini-Flash as an approximation. We tested multiple models at the 32B parameter level, comparing their summed log-probabilities over examples from the dataset and find that they assign similar likelihood (albeit giving different top-k next token predictions) and settled simply using the model we start from as the reference model such that $q, \pi_\mathrm{ref}$ and $\pi_\theta$ are all close at the beginning of training.

\section{Compute resources and infrastructure}
\paragraph{LLM training} All LLM experiments are performed using a single machine with 8 H100 NVIDIA GPUs and 192GB of host ram. 
Training is performed using pytorch \citep{pytorch} and the HuggingFace accelerate library \citep{accelerate}. Training completes within ~18 hours. 

\paragraph{Continuous control experiments} For training models on D4RL~\citep{fu2020d4rl} we use a single machine with an L40 NVIDIA GPU. Though the models here are small and could be trained on any consumer GPU. We use a batch size of 256 where each element consists of a full trajectory (thus we are able to compute normalized importance weights on the fly). Training time depends on the task and accelerator used (we estimate 30-60 minutes per task).

\section{Dataset curation for control tasks}
\label{Appendix:stratefied_data}
The creation of curated datasets is described in Algorithm \ref{alg:curate}.
\begin{table}[t]
\centering
\scalebox{0.78}{
\begin{tabular}{ l|| c  c  c  c  c   c  c }
    \centering
            & Learning Rate & Warmup Steps & Total Steps & $\alpha$ (EMA) & $k / |\tau|$ & percentile cutoffs & normalized weights\\
    \midrule
    \multicolumn{8}{c}{\textbf{Medium Replay (v2)}} \\ \midrule
    halfcheetah  &  4e-5 & 300 & 4500 & 0.995 & 1. & [90, 95, 98]&  False\\
    hopper & 1e-4 & 300 & 4500 & 0.99 & 1. & [90, 95, 98] &  Yes\\
    walker2d &  4e-5 & 300 & 2000& 0.9 & 10. & [90, 95, 98] & Yes\\
    \midrule
    \multicolumn{8}{c}{\textbf{Expert (v2)}} \\ \midrule
    halfcheetah  & 4e-5 & 300 & 4500 & 0.995 & 0.8 & [90, 95, 98], & No \\
    hopper & 4e-4 & 300 & 4500 & 0.995 & 0.8 & [90, 95, 98] & No\\
    walker2d & 4e-5 & 300 & 4500 & 0.995 & 0.8 & [90, 95, 98], & No  
\end{tabular}}
\caption{Hyperparameters for \ourmethod(Q) on the D4RL suite of tasks.}
\label{tab:hyper}
\end{table}

\begin{algorithm}
\caption{Dataset curation for control tasks$\mathcal{D}^+_{Q}$}\label{alg:curate}
\begin{algorithmic}
\Require $\mathcal{D} = \lbrace (\tau^1, r_1), \cdots, (\tau^n, r_n) \rbrace$, $Q$, $\mathcal{D}_S = \emptyset$, percentage cutoffs=$[c_1, \cdots, c_N]$
\Ensure $p_a < p_z \leq 100$
\For{$i=$ \tt{np.arange}(N)}
\State $a_i \gets$ \tt{np.percentile}$(\mathcal{D}, c_i)$
\State $\mathcal{D}_{+} = \lbrace(\tau^i, r_i) : (\tau^i, r_i) \in \mathcal{D}, ~r_i > a_i \rbrace$
\State$\mathcal{D}_{S}\gets\mathcal{D}_S  \cup \mathcal{D}_{+}$
\EndFor
\end{algorithmic}
\end{algorithm}

\section{Additional details for training on D4RL control tasks}
\label{Appendix:Hyper}
For all training runs we use a linear warm-up for the learning rate followed by cosine annealing~\citep{loshchilov2017sgdr} we train with Adam~\citep{KingmaB14}. We also use a three layer MLP with 256 as hidden dimension. We first train a BC policy for 10000 steps with learning rate $1e-3$ and batch size 32. This policy is used as initialization for $\pi_\theta$. For subsequent fine tuning, we use batch size 256 for all tasks. For SFT(Q), we use learning rate 4e-5, warmup steps 300, total steps 4500, percentile cutoffs [90,95, 98], apart from the hopper-medium-expert-v2 where we use learning rate 4e-4. Hyperparameters for \ourmethod(Q) can be found in \ref{tab:hyper}, we note that we first divide 

\section{Ablations with Budget Forcing}\label{Appendix:budget}
\begin{figure}[htb!]
    \centering    \includegraphics[width=0.98\linewidth]{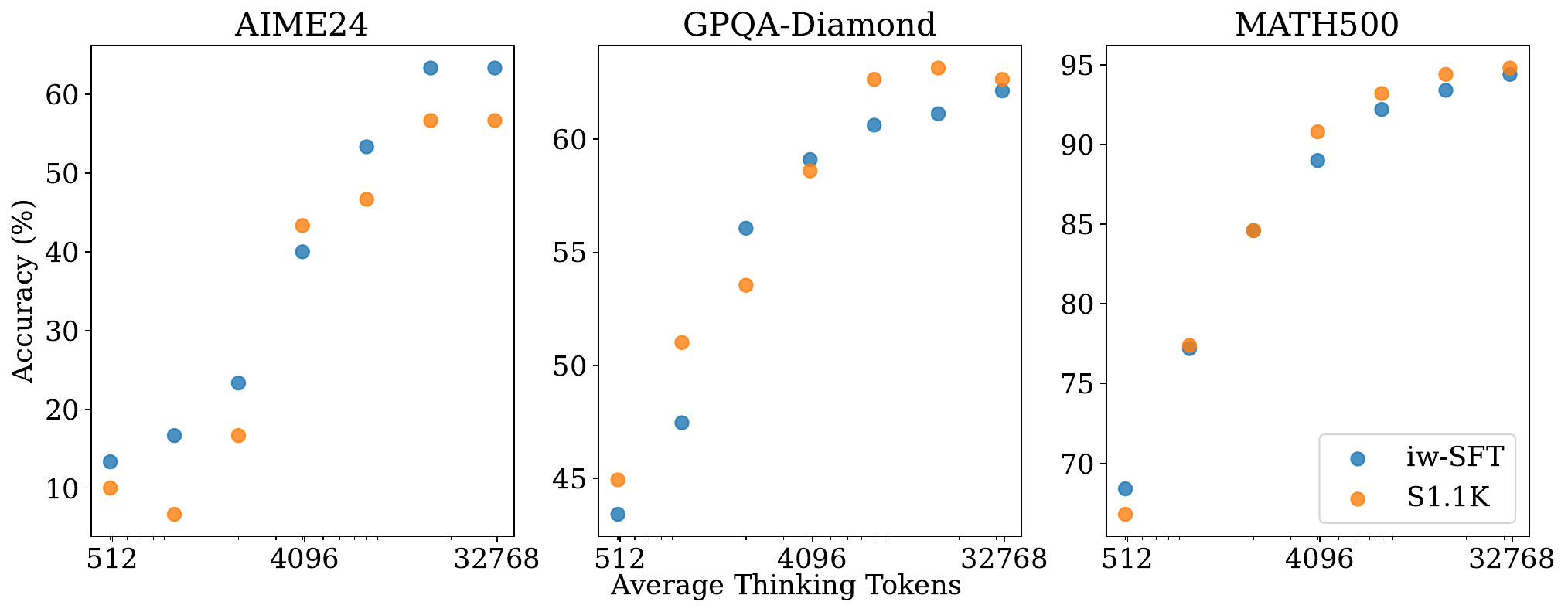}
    \caption{\ourmethod{} shown in blue compared to budged forcing results from \citet{muennighoff2025s1} shown in orange. The x-axis is average thinking tokens and the y-axis corresponds to accuracy on the respective benchmarks. Note that the best result for our method is achieved without explicitly forcing a specific thinking time (cf. main paper, not shown here).}
    \label{fig:budget}
    \vspace{-2em}
\end{figure}

We perform an additoinal experiment to evaluate whether budget forcing as in \citet{muennighoff2025s1} further improves \ourmethod\. The results are shown in Figure~\ref{fig:budget}. As can be seen, when we control the exact amount of thinking tokens and slowly increase it (forcing longer thinking times by injecting "Wait" as proposed in \citet{muennighoff2025s1}) we do not observe an improvement upon the result with no forcing (see Table~\ref{tab:perf}, not included in figure). We note that the graph here was created using an earlier training run with slightly different settings (e.g. for the importance weight truncation) we did not re-run with all token budgets for the latest model due to the additional costs incurred.

\section{Code}
The code to reproduce our results can be found on github. For LLM training the code is available under \url{https://github.com/emmyqin/iw_sft}. The checkpoint is avaliable at huggingface under \url{https://huggingface.co/ChongliQin/iw-SFT-32B}.  

The code for control experiments can also be found on github under \url{https://github.com/emmyqin/iw_sft_control}.

\subsection{Expanded D4RL results}
Expanded Offline RL results for control suite environments from the D4RL suite of tasks using our method are given in Table \ref{tab:finetuning_expand}.

\begin{table}[t]
\caption{Expanded Offline RL results for control suite environments from the D4RL suite of tasks using our method. We show standard deviations for 5 independent runs (different random initializations of the network weights) for our method. We assume unit standard deviations for references (since they were not reported in the literature).}
\label{tab:finetuning_expand}
\centering
\scalebox{0.9}{
\begin{tabular}{ l|| c  c  c  c   c  || c   c  c}
    \centering
            & BC & BC & AWAC & TD3 & IQL & SFT & SFT(Q) & \ourmethod(Q) \\
    Dataset/ Method &    & (10\%) &  & +BC &        &     &        &               \\
    \midrule
    \multicolumn{9}{c}{\textbf{Medium Replay (v2)}} \\ \midrule
    halfcheetah  & 36.6 & 40.6 & 40.5 & \textbf{44.6} &  \textbf{44.2} & 35.1 $\pm$ 4.4 & 39.3 $\pm$ 4.2 & 40.9 $\pm$ 2.6 \\
    hopper & 18.1 & 75.9 & 37.2 & 60.9 & \textbf{94.7} & 79.0 $\pm$ 6.1 & 84.9 $\pm$ 2.9 & 85.0 $\pm$ 4.1 \\
    walker2d & 26.0 & 62.5 & 27.0 & \textbf{81.8} & 73.9 & 58.8 $\pm$ 7.3 & 66.2 $\pm$ 5.5  & 75.8 $\pm$ 2.4 \\
    \midrule
    \multicolumn{9}{c}{\textbf{Expert (v2)}} \\ \midrule
    halfcheetah & 55.2 & \textbf{92.9} & 42.8 & \textbf{90.7} & 86.7 & \textbf{92.3}  $\pm$ 2.7 & \textbf{92.3} $\pm$ 2.9 & \textbf{92.6} $\pm$ 2.7\\
    hopper & 52.5 & \textbf{110.9} & 55.8 & 98.0 & 91.5 & \textbf{111.2} $\pm$ 4.0 & \textbf{109.5} $\pm$ 5.1 & \textbf{111.1} $\pm$ 4.3 \\
    walker2d & \textbf{107.5} & \textbf{109.0} & 74.5 & \textbf{110.1} & \textbf{109.6} & \textbf{109.1} $\pm$ 4.1 & \textbf{109.0} $\pm$ 3.4 & \textbf{109.0} $\pm$ 3.3\\
    \midrule
    \multicolumn{9}{c}{\textbf{Ant-maze (v0)}} \\ \midrule
    umaze & 54.6  & 62.8 & 56.7 & \textbf{87.6} & \textbf{86.7} & 59.7 $\pm$ 9.1 & 60.7 $\pm$ 8.6 & 63.1 $\pm$ 10.5 \\
\end{tabular}}
\end{table}

\end{document}